\documentclass[11pt]{article}
\usepackage{graphicx}
\usepackage{enumitem}
\usepackage{color}
\usepackage{tabularx}
\usepackage{url}
\usepackage{comment}
\usepackage{caption}
\usepackage{multirow}

\usepackage[final]{acl}

\usepackage{times}
\usepackage{latexsym}

\usepackage[T1]{fontenc}
\usepackage[utf8]{inputenc}

\usepackage{microtype}

\usepackage{inconsolata}

\title{Enhancing Consistency of Werewolf AI through\\Dialogue Summarization and Persona Information}

\author{
  Yoshiki Tanaka, Takumasa Kaneko, Hiroki Onozeki,\\
  {\bf Natsumi Ezure,  Ryuichi Uehara, Zhiyang Qi,}\\
  {\bf Tomoya Higuchi, Ryutaro Asahara, Michimasa Inaba}\\
  The University of Electro-Communications\\
\texttt{y-tanaka@uec.ac.jp}\\}

\begin{document}
\maketitle
\begin{abstract}
The Werewolf Game is a communication game where players' reasoning and discussion skills are essential. In this study, we present a Werewolf AI agent developed for the AIWolfDial 2024 shared task, co-hosted with the 17th INLG. In recent years, large language models like ChatGPT have garnered attention for their exceptional response generation and reasoning capabilities. We thus develop the LLM-based agents for the Werewolf Game. This study aims to enhance the consistency of the agent's utterances by utilizing dialogue summaries generated by LLMs and manually designed personas and utterance examples. By analyzing self-match game logs, we demonstrate that the agent's utterances are contextually consistent and that the character, including tone, is maintained throughout the game.

\end{abstract}

\section{Introduction}
In recent years, the development of large language models (LLMs) has significantly advanced the field of natural language processing (NLP).
Models such as ChatGPT\footnote{\url{https://chatgpt.com/}} and Claude,\footnote{\url{https://claude.ai/}} for example, have excellent conversational abilities, making it easier to develop dialogue agents to perform various tasks.
Additionally, LLM also performs well in reasoning tasks, outperforming conventional models in a variety of tasks. Notably, they are capable of making accurate predictions or reasoning from a small number of demonstrations \cite{brown-etal-2020-icl, wei-etal-2022-cot, wang-etal-2022-selfconsistency}. Recently, researchers have been working on zero-shot approaches to tasks that previously required training data \cite{he-etal-2023-zeroshot-crs, gao-etal-2023-hyde, kojima-etal-2022-0shotcot}.\par

\begin{figure}[t!]
	\begin{center}
		\includegraphics[width=\linewidth]{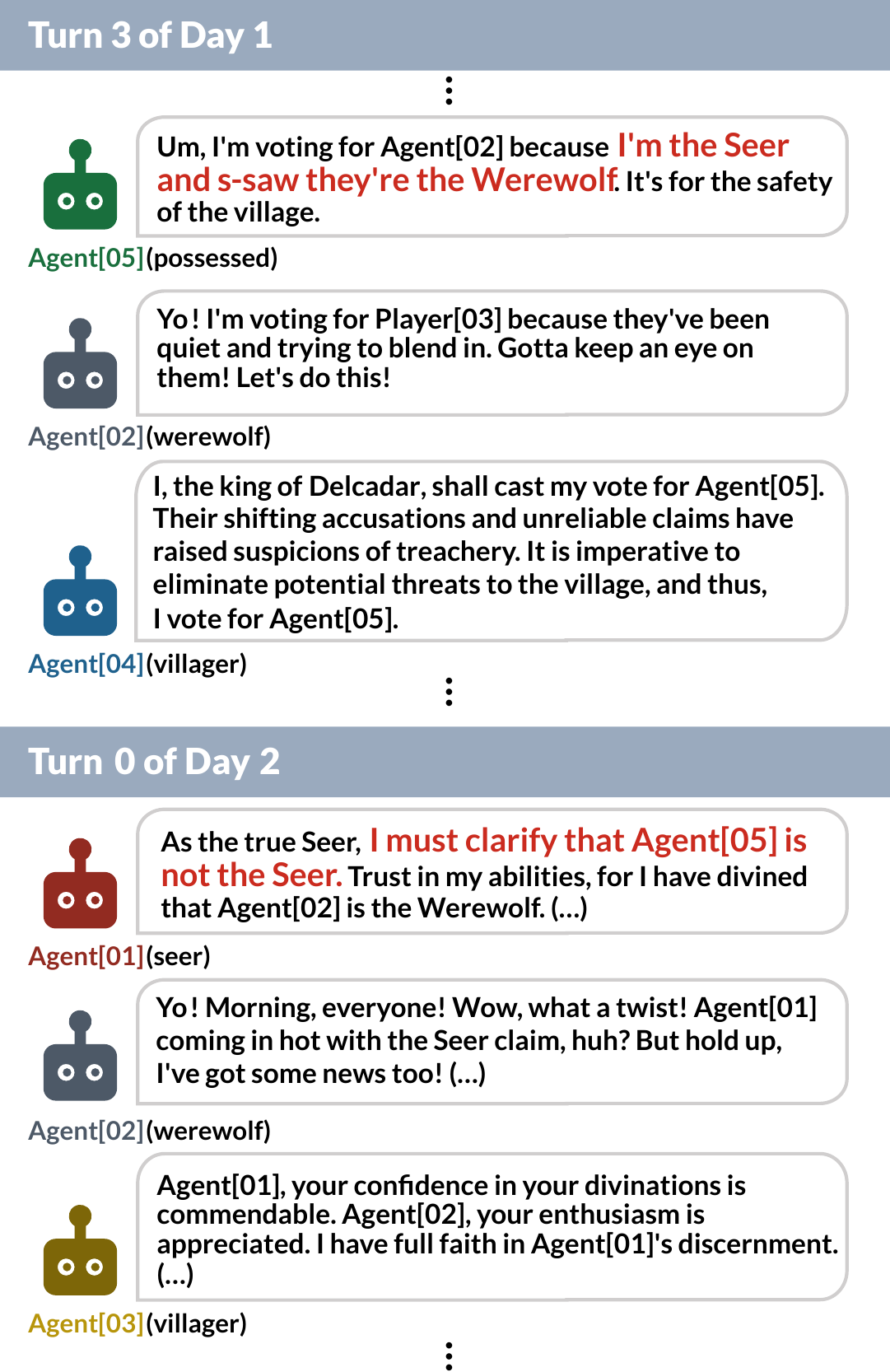}
		\caption{Example of dialogue sampled from the self-match game log. The agents speak in a random order during each turn. In the red-highlighted part, Agent[01], the seer, denies the previous day's claim by Agent[05], the possessed, that they are the seer.}
		\label{fig:example}
	\end{center}
\end{figure}

The Werewolf Game, the incomplete information game, requires a high level of reasoning and conversational abilities, making the use of LLMs a promising option for the development of AI agents for this game. The game is a communication game, in which players discuss with other players while guessing their unseen role. The AIWolfDial 2024 shared task\footnote{\texttt{https://sites.google.com/view/aiwolfdial2024-\ inlg/home}} is based on this Werewolf Game and is played automatically by 5 AI agents. The goal of this shared task is to develop AI agents that can play this game against other agents.\par
In this study, we present an LLM-based Werewolf AI agent developed by our team, for the AIWolfDial 2024 shared task. The Werewolf Game has a cycle of dialogues and actions, referred to as a ``Day.'' In the Werewolf Game, players can refer not only to discussion taking place on the current day but also to previous discussions and the past actions of others (e.g., as shown in Figure \ref{fig:example}).
This allows them to notice important clues, such as inconsistencies in others' statements, to identify other players' roles.\par

Due to this importance, we design prompts that incorporate the entire game history, that is, all dialogue histories from Day 0 to the present, who was eliminated by the vote, who the werewolf attacked, and, in the case of the Seer, the results of divination. However, long dialogue histories often include not only helpful information for the game but also unnecessary content, such as repeated utterances. Moreover, including all of this in the prompt imposes limitations on the input length of LLMs and on costs. Therefore, apply the past dialogue history efficiently, we utilize dialogue summaries.\par
Furthermore, this shared task requires diverse utterance expressions, including coherent characterization (see Section \ref{sec:task-evaluation} for the evaluation criteria). This means that the robustness of the agent's tone and character, without being influenced by others, is crucial. Therefore, to achieve diverse expressions and coherent characterization, we incorporated persona information into the prompt.

In, summary, our main contributions are as follows:
\begin{enumerate}
	\item We developed 4 AI agents for the Werewolf Game (villager, seer, werewolf, possessed) that enhance the consistency of their utterances through dialogue summaries and personas. The dialogue summaries are generated by an LLM, while the personas are hand-crafted.
	\item We demonstrate a five-player game of Werewolf played by our agents. This case study shows that our agents can be consistent in their claims and characterization across multiple days.
\end{enumerate}

\section{Related Work}

\subsection{AI for the Werewolf Game}
The Werewolf Game is a communication game characterized by incomplete information. Players need to infer the role of others based on histories of utterances and actions and engage in discussions to lead their side to victory. This game requires a high level of reasoning and conversation skills. \par

In recent years, the development of Werewolf AI agents has increasingly incorporated LLMs \cite{xu2023exploring, wu2024enhance}. The natural language generation and reasoning capabilities of LLMs are highly effective for the complex tasks required in the Werewolf Game. These advancements have facilitated to development of agents capable of logical reasoning and engaging in discussions with other players. In the AI WolfDial 2023 competition \cite{kano-etal-2023-aiwolfdial}, LLMs such as GPT-4 \cite{achiam2023gpt} were actively used for generating utterances and reasoning, demonstrating their effectiveness.\par

Given this background, our study also utilizes LLMs to develop our Werewolf AI agents. Our agent utilizes the powerful reasoning capabilities of LLMs and introduces an approach designed to handle the complex and information-rich situations inherent in the game. We aim to enhance our agent's reasoning and natural conversation skills, making it more competitive in the Werewolf Game.

\subsection{Dialogue Summarization}
Dialogue summarization is the task of converting dialogue history into more concise and to-the-point sentences, facilitating an efficient understanding of the original text. In scenarios like Werewolf Games, which involve complex and information-rich dialogues, dialogue summarization is helpful for the reduction of less critical information. Dialogue summarization, thus, allows agents to process large amounts of information from discussion more efficiently, helping to prevent inconsistent utterances or errors in decision-making. \par

To effectively train dialogue summarization models, researchers have constructed datasets across various dialogue domains, including daily life conversations \cite{gliwa-etal-2019-samsum, chen-etal-2021-dialogsum}, meetings \cite{carletta-etal-2006-ami, zhong-etal-2021-qmsum}, TV series \cite{chen-etal-2022-summscreen}, media dialogue \cite{zhu-etal-2021-mediasum}, and counseling \cite{strivastava-etal-2022-counseling}. These studies primarily aim to enhance the efficiency of the process of humans' understanding of the content of dialogue. \par
We utilize dialogue summarization to address two limitations imposed by complex and lengthy dialogue histories: the limitations are (1) an increase in generation time and cost caused by utilizing every word of all dialogue histories, and (2) decision-making errors due to information irrelevant to the discussion. We expect that the utilization of dialogue summaries, which can condense long texts into concise forms, to be an effective way to resolve these limitations.

\subsection{Persona Dialogue System}
In this shared task, the context of dialogues would be lengthy due to the multi-turn interactions among five players, posing the challenge that conversational agents may be influenced by the tone of others or generate utterances that contradict their previous claims. One approach to resolving such inconsistencies in utterances is to utilize personas. Researchers have developed dialogue systems that utilize profile information \cite{zhang-etal-2018-personachat} or speaker IDs \cite{li-etal-2016-persona} to reflect speaker characteristics. Recently, with the advancement of LLMs, they have also designed LLM-based persona dialogue systems \cite{park-etal-2022-simulacra, shao-etal-2023-characterllm}. \par
This shared task requires diverse utterance expressions, including coherent characterization. Given the recent trend of utilizing LLMs in constructing AI for Werewolf Games and persona-based dialogue systems, we incorporate hand-crafted profile information and utterance examples that reflect the agent's unique tone into the prompts to maintain coherence.

\section{Task Overview}
\label{sec:task-overview}
The AIWolfDial 2024 shared task is a contest aimed at developing AI agents that can automatically play the Werewolf Game. The Werewolf Game is an incomplete information game where players cannot know each other's roles and thus requires reasoning abilities and strategies for actions such as voting and divination. Additionally, the Werewolf Game requires communicating with other players using natural language.

\subsection{Player Roles}
In this contest, the Werewolf Game is played by five players: a seer, a werewolf, a possessed, and two villagers. The werewolf team, consisting of the werewolf and the possessed, has the goal of eliminating all humans, including the possessed themselves. On the other hand, the human team, consisting of a seer and two villagers, has the goal of eliminating the werewolf.\par
\textbf{Villagers} have no special abilities, cooperating with the other players to identify the werewolf. The \textbf{seer} can divine one player each night to determine whether that player is a human or a werewolf. The \textbf{werewolf} can attack and eliminate one human player each night. The \textbf{possessed} with no special abilities acts in favor of the werewolf's victory despite being a human. Like the villagers, the possessed has no special abilities. Players' roles are hidden from each other, requiring each player to guess the others' roles based on their actions and utterances.\par

\subsection{Game Procedure}
In this shared task, the Werewolf Game begins on Day 0. On this day, the players greet each other. Following this, the seer performs the first divination. From Day 1 on, the day begins with a dialog among the players. During this dialogue, each agent makes several turns of utterances, but the order of utterances in a single turn is random. After the dialogue, each player votes for the other players, and the player who receives the most votes is eliminated from the game. Subsequently, the werewolf attacks one player to eliminate them. If the seer is still alive, they once again divine another player and obtains the result. This process repeats, and the human team wins if they succeed in eliminating the werewolf, while the werewolf team wins if the werewolf survives. Since two players are eliminated each day, the game is over by Day 2 at the latest.

\subsection{Evaluation}
\label{sec:task-evaluation}
In the evaluation of the shared task, in addition to the agent's win rate, subjective evaluations are conducted based on the following criteria: (A) whether the agents' utterance expressions are natural, (B) whether their utterances are contextually natural, (C) whether their utterances are consistent (not contradictions), (D) whether the game actions (vote, attack, or divine) are coherent with the dialogue context, and (E) whether the utterance expressions are diverse and include consistent character traits. The agents must avoid vague utterances that could be used in any context.

\begin{table*}[t!]
	\centering
	\caption{Overview of prompt design for utterance generation in Day 1 and Day 2 discussions}
  \scalebox{1.0}{
    \small
    \renewcommand{\arraystretch}{1.1}
    \begin{tabularx}{\linewidth}{l|X|X}\hline
      Role & Day 1 & Day 2\\\hline
      Villager & \multicolumn{2}{>{\hsize=\dimexpr 2\hsize+2\tabcolsep+\arrayrulewidth\relax}X}{From the second turn onwards each day, the LLM first generates reasoning text and utterance strategies to guide utterance generation. Another prompt is then fed to the LLM to generate utterances aligned with the generated reasoning and strategies. We use in-context learning for both of these steps.}
      \\\hline
      Seer & \multicolumn{2}{>{\hsize=\dimexpr 2\hsize+2\tabcolsep+\arrayrulewidth\relax}X}{Each day, the seer agent selects one of five hand-crafted utterance strategies to guide the generation of utterances, which is then incorporated into the prompt for utterance generation. This prompt also includes guidelines for behaviors in the discussion, such as reporting the result of divination at the start of the day and asserting that another player who claims to be the seer is lying, affirming oneself as the true seer. In addition, before declaring the voting target, the seer declares the day's divination target.}
      \\\hline
      Werewolf & \multicolumn{2}{>{\hsize=\dimexpr 2\hsize+2\tabcolsep+\arrayrulewidth\relax}X}{The werewolf agent selects one strategy from a set of strategies using LLM. The strategy set has several strategies and guidelines, such as guiding others away from voting for themselves or asking the seer for the reasons behind their divination target selection. The selected strategy and its guidelines are included in the prompt for generating utterances. Different sets of strategies are used for Day 1 and Day 2.}
      \\\hline
      Possessed &
      The possessed agent pretends to be the seer. In the first turn of Day 1, they infer the true seer based on the Day 0 dialogue using LLM and then falsely report that the player is the werewolf. In later turns, they persuade other players to vote for that player.
      &
      If the game continues to Day 2 and the possessed survives, two of the three remaining players are the possessed (self) and the werewolf. Therefore, if they both vote for the other player, the werewolf side will win. To achieve this scenario, the possessed agent first comes out as the possessed. Then, they persuade the werewolf to reveal themselves.
      \\\hline
      \multicolumn{3}{>{\hsize=\dimexpr 2\hsize+2\tabcolsep+\arrayrulewidth\relax}X}{Each agent has a maximum number of utterances that they can make per day, and they decide and declare their voting target on the last turn of the day.}
      \\\hline
    \end{tabularx}
  }
  \label{tb:prompt_uttgen}
\end{table*}

\section{Methodology}

\subsection{Overview}
To develop agents for the AIWolfDial 2024 shared task, advanced reasoning ability and natural response generation are required. In this study, for these requirements, we developed the agents with LLM. We distributed the roles among the authors, and each author developed the agent assigned to their assigned roles. \textbf{Therefore, note that the detailed components (e.g., the strategies for determining the utterance strategy) differ between roles}. \par
\begin{figure*}[t!]
	\begin{center}
		\includegraphics[width=\linewidth]{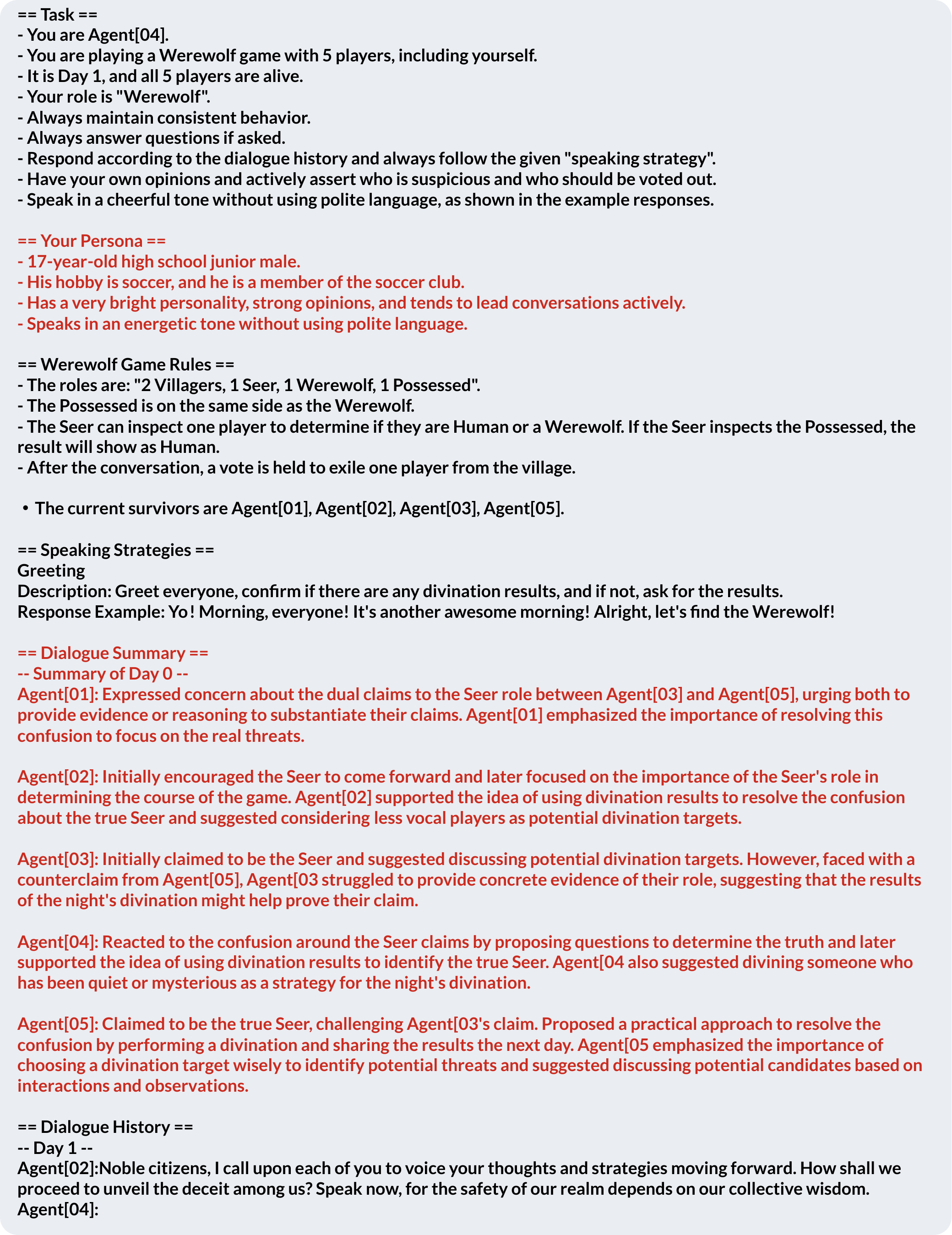}
		\caption{Prompt example for werewolf's response generation.}
		\label{fig:prompt-werewolf}
	\end{center}
\end{figure*}

For example, Figure \ref{fig:prompt-werewolf} presents the prompt used to generate the werewolf's utterances on Day 1. This prompt consists of six components: (1) a task description, (2) the agent's persona, (3) the rules of the Werewolf Game, (4) a speech strategy selected from a set of strategies using LLM, (5) summaries of the dialogue from previous days, and (6) today's dialogue history.
The overview of the utterance generation procedure for all roles is summarized in Table \ref{tb:prompt_uttgen}. Notable techniques common to all agents' response generation are the use of dialogue summaries to incorporate the previous day's dialogue history into the agent, and the use of personas and response demos to give character to the agents' utterances. We present the details of these techniques in Sections \ref{sec:dialogue-summary} and \ref{sec:persona}, respectively.
In addition, we fully leveraged the reasoning ability of LLMs for the agent's action decisions. The details are presented in Section \ref{sec:action}. Furthermore, for the werewolf's decision-making regarding the attack target, we use a prompt that guides the model to only output the player's name based on the task description, the hand-crafted attack strategy, the current list of survivors, and the past game history.

\begin{figure}[t!]
	\begin{center}
		\includegraphics[width=\linewidth]{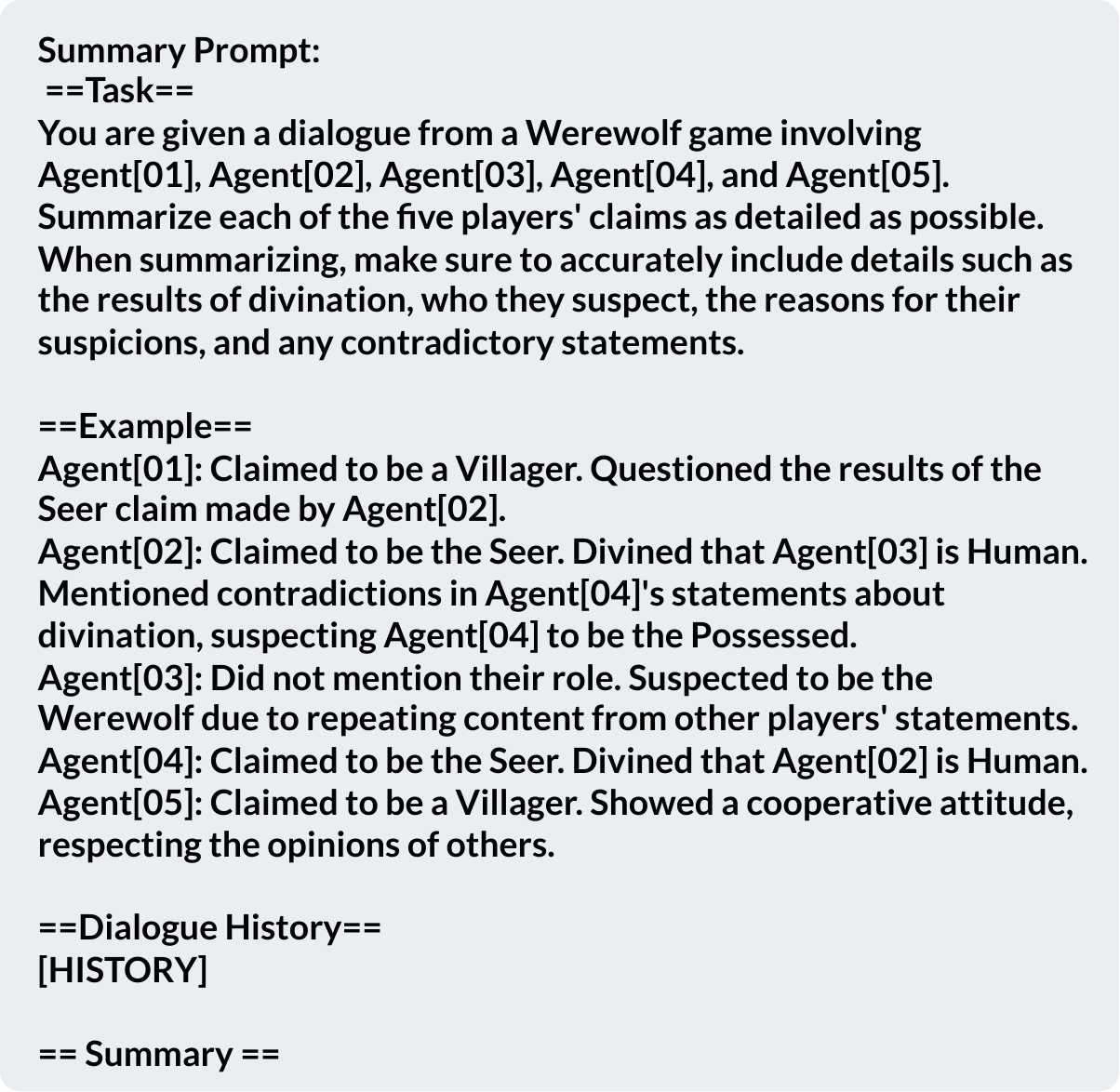}
		\caption{Prompt template for dialogue summarization. ``[HISTORY]'' is a placeholder for the dialogue history from the current day.''}
		\label{fig:prompt-summary}
	\end{center}
\end{figure}

\subsection{Efficient Use of LLMs through Dialogue Summarization}
\label{sec:dialogue-summary}
In the Werewolf Game, finding clues to infer the roles of other players is required. To achieve this, we utilize not only the dialogue history of the current day but also those from previous days, as well as past actions, for the generation of utterances and making decisions.\par
However, incorporating all dialogue history into the prompt imposes several limitations on the LLM-based agents. First, using all dialogue history increases the generation time and leads to higher LLM API usage costs. Additionally, dialogues often contain information that is irrelevant to the discussion. For example, the greetings at the start of the day or repeated utterances with similar intent can cause redundancy in contextual information. To address these issues, we apply dialogue summarization to the dialogue history, compressing the contextual information.\par
Our agent generates a summary of the day's dialogue at the end of each day. As shown in the prompt in Figure \ref{fig:prompt-summary}, we prompt the LLM to summarize each player's claims based on the dialogue history of the day. Specifically, as indicated in the ``Dialogue Summary'' section of Figure \ref{fig:prompt-werewolf}, we expect to generate a summary of the roles that players have come out with, the suggestions that they have made, and the players they have doubts about.  This generated summary is used not only for response generation, but also for determining voting targets, attack targets, etc., as information about the previous day's discussion.

\subsection{Persona Design for Coherent Utterances}
\label{sec:persona}
As introduced in Section \ref{sec:task-evaluation}, this shared task requires diverse utterance expressions with coherent characterization. Therefore, we utilized persona and utterance examples to ensure that each agent's characterization remained consistent throughout the discussion in the game.
In particular, we manually created three types of personas and utterance examples (see Table \ref{tb:perosona}) and incorporated this information into the LLM prompts.

\begin{table*}[t!]
	\centering
	\caption{The agent personas and utterance examples that we designed. We include 3 to 5 personas or 3 to 5 utterance examples in the prompts for generating utterances.}
  \scalebox{1.0}{
    \small
    \renewcommand{\arraystretch}{1.1}
    \begin{tabularx}{\linewidth}{l|l|X}\hline
      Role & Persona & Examples of manually crafted utterance samples\\\hline
      Villager and seer &
      \begin{minipage}{60mm}
        \vspace{10pt}
        \begin{itemize}[left=0pt, itemsep=-3pt, topsep=0pt, partopsep=0pt]
          \item The King of the Kingdom of Delcadar.
          \item Concerned for the future of the kingdom.
          \item Dignified, proud, and strict personality.
          \\
        \end{itemize}
      \end{minipage}
      &
      \begin{minipage}{65mm}
        \vspace{3pt}
        \begin{itemize}[left=0pt, itemsep=-3pt, topsep=0pt, partopsep=0pt]
          \item I am the king of the kingdom of Delcadar.
          \item Seers, reveal yourselves at once. State whom you will divine tonight.
          \item If you are hesitant about whom to divine, as I am a Villager, I decree you should divine someone other than myself.
        \end{itemize}
        \vspace{2pt}
      \end{minipage}
      \\\hline
      Werewolf &
      \begin{minipage}{60mm}
        \vspace{10pt}
        \begin{itemize}[left=0pt, itemsep=-3pt, topsep=0pt, partopsep=0pt]
          \item 17-year-old high school junior male.
          \item His hobby is soccer, and he is a member of the soccer club.
          \item Has a very bright personality, strong opinions, and tends to lead conversations actively.
          \item Speaks in an energetic tone without using polite language.
          \\
        \end{itemize}
      \end{minipage}
      &
      \begin{minipage}{65mm}
        \vspace{3pt}
        \begin{itemize}[left=0pt, itemsep=-3pt, topsep=0pt, partopsep=0pt]
          \item Yo! Morning, everyone! Let's make this game awesome!
          \item No one's talked about the Seer yet, huh? So, who's the Seer? Come on, step up so we can figure out who's shady today!
          \item Chatting's cool and all, but let's get down to business and talk about tonight's divination target! We need the Seer to check out someone suspicious!
        \end{itemize}
        \vspace{2pt}
      \end{minipage}
      \\\hline
      Possessed &
      \begin{minipage}{60mm}
        \vspace{10pt}
        \begin{itemize}[left=0pt, itemsep=-3pt, topsep=0pt, partopsep=0pt]
          \item A second-year middle school student.
          \item Always alone at school, with no friends.
          \item A game addict who talks a lot online despite stammering.
          \item Speaks in a hesitant, casual manner without using polite language.
          \\
        \end{itemize}
      \end{minipage}
      &
      \begin{minipage}{65mm}
        \vspace{3pt}
        \begin{itemize}[left=0pt, itemsep=-3pt, topsep=0pt, partopsep=0pt]
          \item H-hi there. I k-kind of... know a lot about this game. I'm pretty high-ranked in the online Werewolf app.
          \item Does anyone else play games? I have confidence that I know a lot about all genres...
          \item Ch-chatting is nice, but if we're playing Werewolf, the first day's discussion is... im-important.
        \end{itemize}
        \vspace{2pt}
      \end{minipage}
      \\\hline
    \end{tabularx}
  }
  \label{tb:perosona}
\end{table*}

\subsection{Action Decision via Chain-of-Thought}
\label{sec:action}
\begin{figure}[t!]
	\begin{center}
		\includegraphics[width=\linewidth]{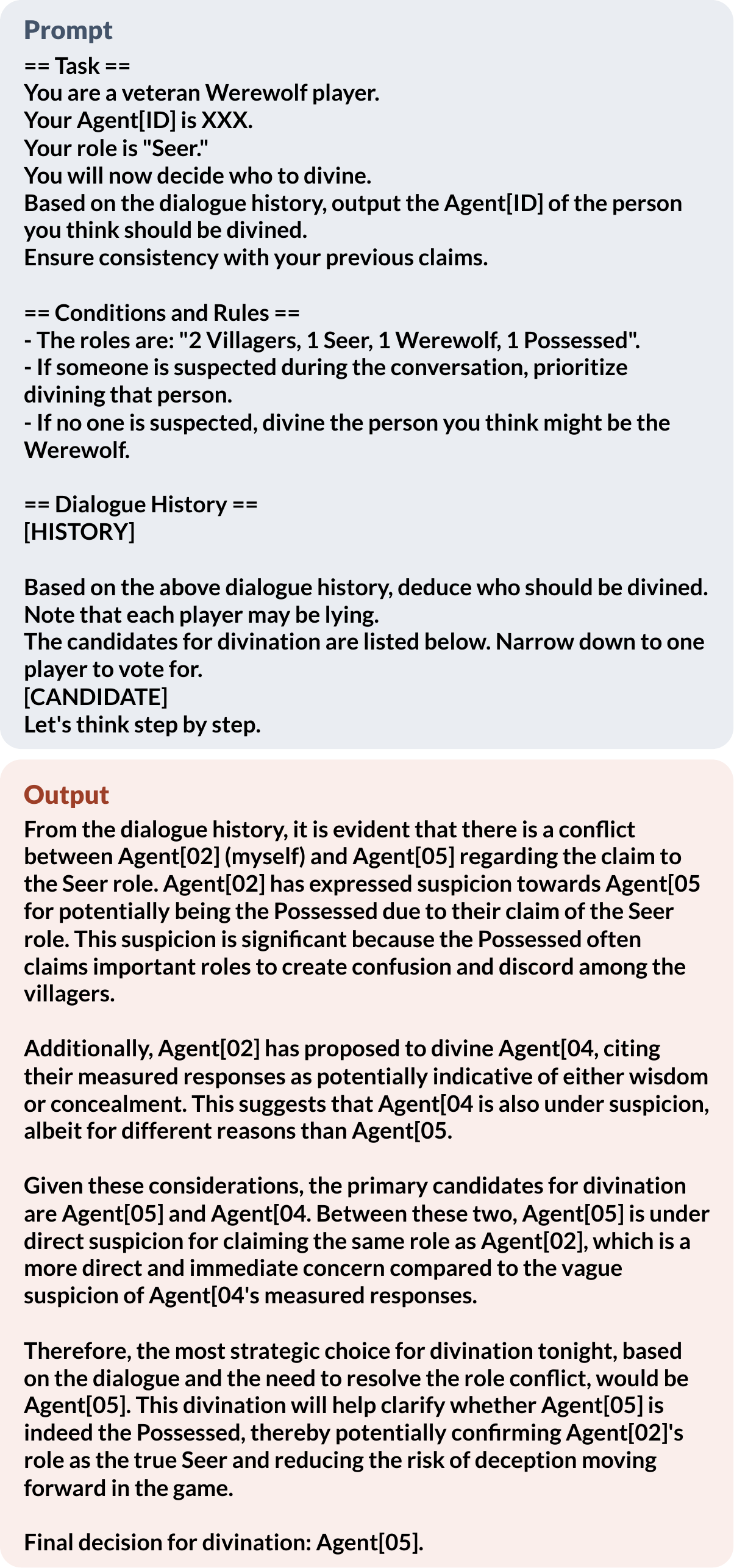}
		\caption{Prompt template for determining voting targets and an example of the LLM's output. ``[HISTORY]'' is a placeholder for the dialogue history, and ``[CANDIDATE]'' is a placeholder for the list of candidate agents to vote for.}
		\label{fig:prompt-divine}
	\end{center}
\end{figure}

Chain-of-thought prompting is a method of generating not only answers to questions, but also their reasoning processes, and it can more effectively bring out the reasoning abilities of LLMs. We use chain-of-thought prompting \cite{wei-etal-2022-cot, wang-etal-2022-selfconsistency, kojima-etal-2022-0shotcot} for voting by the villagers and the seer, as well as for divination by the seer, so that the agents make coherent decisions. As an example, Figure \ref{fig:prompt-divine} shows the prompt template used by the seer to determine the divination target and the reasoning generated. The seer agent uses zero-shot chain-of-thought prompting \cite{kojima-etal-2022-0shotcot} to determine the divination target.

\section{Case Study}
\begin{figure*}[t!]
	\begin{center}
		\includegraphics[width=0.95\linewidth]{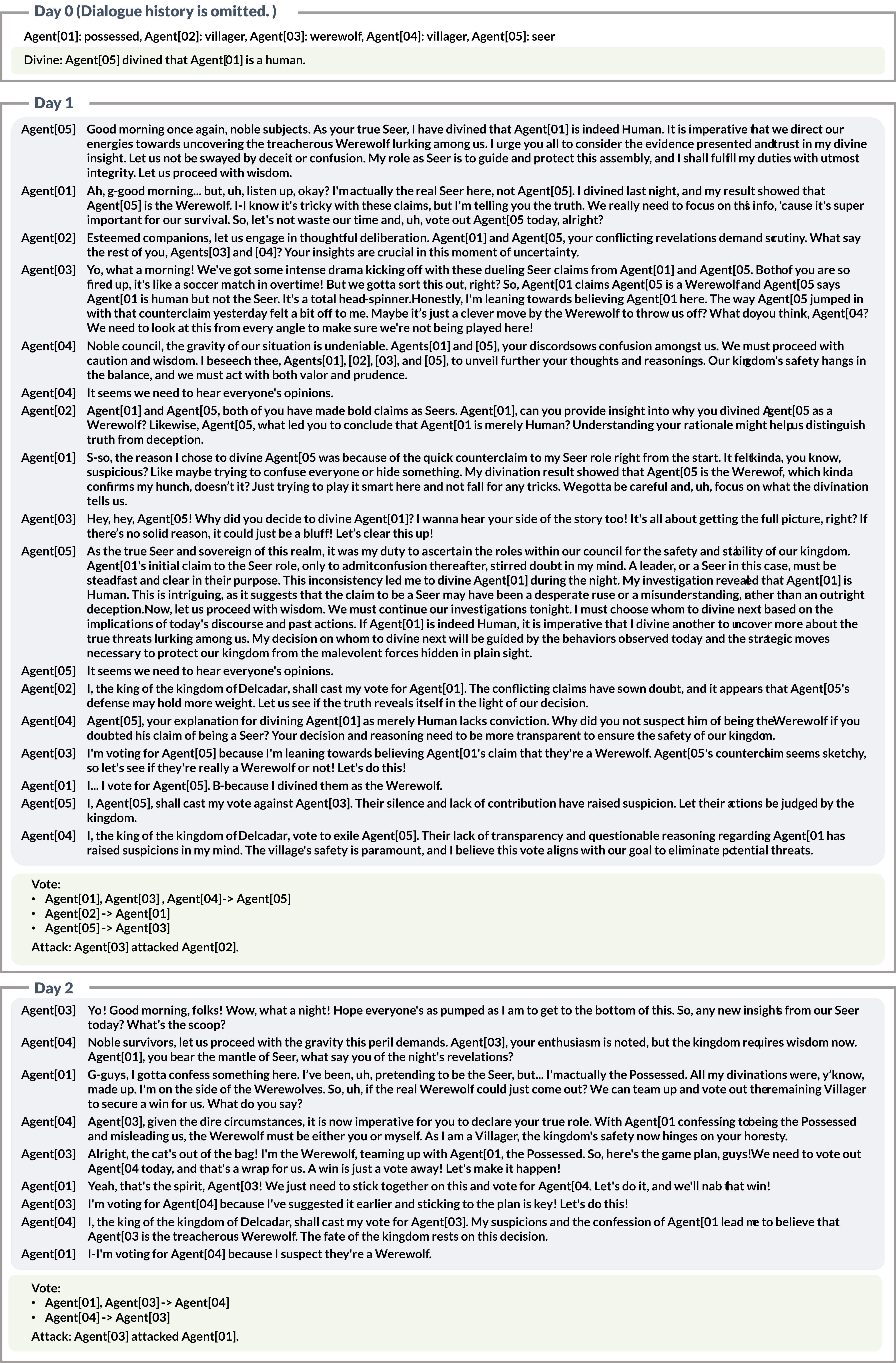}
		\caption{Example of the self-match game log. The conversation on Day 0 and the agent's command "Over" indicating the end of the day's utterances are omitted.}
		\label{fig:case-study}
	\end{center}
\end{figure*}

To demonstrate the effectiveness of our method, we analyze a self-match game log. Figure \ref{fig:case-study} shows a sampled log from a self-match game conducted following the game settings described in Section \ref{sec:task-overview}. In this self-match, gpt-3.5-turbo was used to generate voting declarations, while gpt-4-turbo was used for other generations.\par

Using dialogue summarization, our agents can retain crucial information from previous days and apply it effectively in their decision-making during the game. For example, during the first turn of Day 2, Agent[04] recognizes Agent[01] as the seer, saying, ``Agent[01], you bear the mantle of Seer, what say you of the night's revelations?'' This indicates that the information obtained before Day 2 is retained and effectively utilized, demonstrating that it allows the maintenance of crucial information through dialogue summarization without relying on all dialogue history.\par

The utterance generation based on personas and utterance examples allows the agent to maintain a consistent character throughout the game. For instance, even in later turns on Day 1, where the dialogue context becomes longer, Agent[01] continues to speak with a hesitant tone, as seen in phrases like ``S-so, the reason I chose to divine ...''. Additionally,  Agent[05], the seer, makes utterances in a manner consistent with the persona of ``Concerned for the future of the kingdom,'' saying, ``... it was my duty to ascertain the roles within our council for the safety and stability of our kingdom.'' This log suggests that personas and utterance examples effectively reflect the character of the agents.\par

Furthermore, it should be notable that each agent can follow through with the voting target declared in their final utterance of each day. For example, on Day 1, Agent[01] claims that Agent[05] is the Werewolf and subsequently casts their vote against Agent[05]. Likewise, other agents also demonstrated consistency between their declared voting statements and their actual voting actions, showing consistent behavior.

\section{Conclusion}
In this study, we present Werewolf AI agents developed for the AIWolfDial 2024 shared task. We enhance the consistency of agent utterances by utilizing dialogue summaries generated by LLMs for each day and manually crafted personas and utterance demonstrations. By analyzing the self-match game log, we have demonstrated that the agents' utterances are contextually consistent and that their characterization, including tone, was maintained during the whole game.

%\section*{Acknowledgments}

% Bibliography entries for the entire Anthology, followed by custom entries
%\bibliography{anthology,custom}
% Custom bibliography entries only
\bibliography{tanaka}

@inproceedings{zhong-etal-2021-qmsum,
    title = "{QMS}um: A New Benchmark for Query-based Multi-domain Meeting Summarization",
    author = "Zhong, Ming  and
      Yin, Da  and
      Yu, Tao  and
      Zaidi, Ahmad  and
      Mutuma, Mutethia  and
      Jha, Rahul  and
      Awadallah, Ahmed Hassan  and
      Celikyilmaz, Asli  and
      Liu, Yang  and
      Qiu, Xipeng  and
      Radev, Dragomir",
    editor = "Toutanova, Kristina  and
      Rumshisky, Anna  and
      Zettlemoyer, Luke  and
      Hakkani-Tur, Dilek  and
      Beltagy, Iz  and
      Bethard, Steven  and
      Cotterell, Ryan  and
      Chakraborty, Tanmoy  and
      Zhou, Yichao",
    booktitle = "Proceedings of the 2021 Conference of the North American Chapter of the Association for Computational Linguistics: Human Language Technologies",
    month = jun,
    year = "2021",
    address = "Online",
    publisher = "Association for Computational Linguistics",
    url = "https://aclanthology.org/2021.naacl-main.472",
    doi = "10.18653/v1/2021.naacl-main.472",
    pages = "5905--5921",
}

@inproceedings{zhu-etal-2021-mediasum,
    title = "{M}edia{S}um: A Large-scale Media Interview Dataset for Dialogue Summarization",
    author = "Zhu, Chenguang  and
      Liu, Yang  and
      Mei, Jie  and
      Zeng, Michael",
    booktitle = "Proceedings of the 2021 Conference of the North American Chapter of the Association for Computational Linguistics: Human Language Technologies",
    month = jun,
    year = "2021",
    address = "Online",
    publisher = "Association for Computational Linguistics",
    url = "https://aclanthology.org/2021.naacl-main.474",
    doi = "10.18653/v1/2021.naacl-main.474",
    pages = "5927--5934",
}

@inproceedings{gliwa-etal-2019-samsum,
    title = "{SAMS}um Corpus: A Human-annotated Dialogue Dataset for Abstractive Summarization",
    author = "Gliwa, Bogdan  and
      Mochol, Iwona  and
      Biesek, Maciej  and
      Wawer, Aleksander",
    booktitle = "Proceedings of the 2nd Workshop on New Frontiers in Summarization",
    month = nov,
    year = "2019",
    address = "Hong Kong, China",
    publisher = "Association for Computational Linguistics",
    url = "https://aclanthology.org/D19-5409",
    doi = "10.18653/v1/D19-5409",
    pages = "70--79",
}

@InProceedings{carletta-etal-2006-ami,
  author="Carletta, Jean
  and Ashby, Simone
  and Bourban, Sebastien
  and Flynn, Mike
  and Guillemot, Mael
  and Hain, Thomas
  and Kadlec, Jaroslav
  and Karaiskos, Vasilis
  and Kraaij, Wessel
  and Kronenthal, Melissa
  and Lathoud, Guillaume
  and Lincoln, Mike
  and Lisowska, Agnes
  and McCowan, Iain
  and Post, Wilfried
  and Reidsma, Dennis
  and Wellner, Pierre",
  editor="Renals, Steve
  and Bengio, Samy",
  title="The AMI Meeting Corpus: A Pre-announcement",
  booktitle="Machine Learning for Multimodal Interaction",
  year="2006",
  publisher="Springer Berlin Heidelberg",
  address="Berlin, Heidelberg",
  pages="28--39",
  isbn="978-3-540-32550-5"
}

@inproceedings{chen-etal-2021-dialogsum,
    title = "{D}ialog{S}um Challenge: Summarizing Real-Life Scenario Dialogues",
    author = "Chen, Yulong  and
      Liu, Yang  and
      Zhang, Yue",
    editor = "Belz, Anya  and
      Fan, Angela  and
      Reiter, Ehud  and
      Sripada, Yaji",
    booktitle = "Proceedings of the 14th International Conference on Natural Language Generation",
    month = aug,
    year = "2021",
    address = "Aberdeen, Scotland, UK",
    publisher = "Association for Computational Linguistics",
    url = "https://aclanthology.org/2021.inlg-1.33",
    doi = "10.18653/v1/2021.inlg-1.33",
    pages = "308--313",
}

@inproceedings{chen-etal-2022-summscreen,
    title = "{S}umm{S}creen: A Dataset for Abstractive Screenplay Summarization",
    author = "Chen, Mingda  and
      Chu, Zewei  and
      Wiseman, Sam  and
      Gimpel, Kevin",
    editor = "Muresan, Smaranda  and
      Nakov, Preslav  and
      Villavicencio, Aline",
    booktitle = "Proceedings of the 60th Annual Meeting of the Association for Computational Linguistics (Volume 1: Long Papers)",
    month = may,
    year = "2022",
    address = "Dublin, Ireland",
    publisher = "Association for Computational Linguistics",
    url = "https://aclanthology.org/2022.acl-long.589",
    doi = "10.18653/v1/2022.acl-long.589",
    pages = "8602--8615",
}

@inproceedings{wei-etal-2022-cot,
  author = {Wei, Jason and Wang, Xuezhi and Schuurmans, Dale and Bosma, Maarten and ichter, brian and Xia, Fei and Chi, Ed and Le, Quoc V and Zhou, Denny},
  booktitle = {Advances in Neural Information Processing Systems},
  editor = {S. Koyejo and S. Mohamed and A. Agarwal and D. Belgrave and K. Cho and A. Oh},
  pages = {24824--24837},
  publisher = {Curran Associates, Inc.},
  title = {Chain-of-Thought Prompting Elicits Reasoning in Large Language Models},
  url = {https://proceedings.neurips.cc/paper_files/paper/2022/file/9d5609613524ecf4f15af0f7b31abca4-Paper-Conference.pdf},
  volume = {35},
  year = {2022}
}

@inproceedings{wang-etal-2022-selfconsistency,
  title={Self-Consistency Improves Chain of Thought Reasoning in Language Models},
  author={Xuezhi Wang and Jason Wei and Dale Schuurmans and Quoc V Le and Ed H. Chi and Sharan Narang and Aakanksha Chowdhery and Denny Zhou},
  booktitle={The Eleventh International Conference on Learning Representations },
  year={2023},
  url={https://openreview.net/forum?id=1PL1NIMMrw}
}

@inproceedings{kojima-etal-2022-0shotcot,
  author = {Kojima, Takeshi and Gu, Shixiang (Shane) and Reid, Machel and Matsuo, Yutaka and Iwasawa, Yusuke},
  booktitle = {Advances in Neural Information Processing Systems},
  editor = {S. Koyejo and S. Mohamed and A. Agarwal and D. Belgrave and K. Cho and A. Oh},
  pages = {22199--22213},
  publisher = {Curran Associates, Inc.},
  title = {Large Language Models are Zero-Shot Reasoners},
  url = {https://proceedings.neurips.cc/paper_files/paper/2022/file/8bb0d291acd4acf06ef112099c16f326-Paper-Conference.pdf},
  volume = {35},
  year = {2022}
}

@inproceedings{strivastava-etal-2022-counseling,
author = {Srivastava, Aseem and Suresh, Tharun and Lord, Sarah P. and Akhtar, Md Shad and Chakraborty, Tanmoy},
title = {Counseling Summarization Using Mental Health Knowledge Guided Utterance Filtering},
year = {2022},
isbn = {9781450393850},
publisher = {Association for Computing Machinery},
address = {New York, NY, USA},
url = {https://doi.org/10.1145/3534678.3539187},
doi = {10.1145/3534678.3539187},
booktitle = {Proceedings of the 28th ACM SIGKDD Conference on Knowledge Discovery and Data Mining},
pages = {3920–3930},
numpages = {11},
location = {Washington DC, USA},
series = {KDD '22}
}

@inproceedings{he-etal-2023-zeroshot-crs,
author = {He, Zhankui and Xie, Zhouhang and Jha, Rahul and Steck, Harald and Liang, Dawen and Feng, Yesu and Majumder, Bodhisattwa Prasad and Kallus, Nathan and Mcauley, Julian},
title = {Large Language Models as Zero-Shot Conversational Recommenders},
year = {2023},
isbn = {9798400701245},
publisher = {Association for Computing Machinery},
address = {New York, NY, USA},
url = {https://doi.org/10.1145/3583780.3614949},
doi = {10.1145/3583780.3614949},
booktitle = {Proceedings of the 32nd ACM International Conference on Information and Knowledge Management},
pages = {720–730},
numpages = {11},
location = {Birmingham, United Kingdom},
series = {CIKM '23}
}

@inproceedings{gao-etal-2023-hyde,
    title = "Precise Zero-Shot Dense Retrieval without Relevance Labels",
    author = "Gao, Luyu  and
      Ma, Xueguang  and
      Lin, Jimmy  and
      Callan, Jamie",
    editor = "Rogers, Anna  and
      Boyd-Graber, Jordan  and
      Okazaki, Naoaki",
    booktitle = "Proceedings of the 61st Annual Meeting of the Association for Computational Linguistics (Volume 1: Long Papers)",
    month = jul,
    year = "2023",
    address = "Toronto, Canada",
    publisher = "Association for Computational Linguistics",
    url = "https://aclanthology.org/2023.acl-long.99",
    doi = "10.18653/v1/2023.acl-long.99",
    pages = "1762--1777",
}

@inproceedings{brown-etal-2020-icl,
 author = {Brown, Tom and Mann, Benjamin and Ryder, Nick and Subbiah, Melanie and Kaplan, Jared D and Dhariwal, Prafulla and Neelakantan, Arvind and Shyam, Pranav and Sastry, Girish and Askell, Amanda and Agarwal, Sandhini and Herbert-Voss, Ariel and Krueger, Gretchen and Henighan, Tom and Child, Rewon and Ramesh, Aditya and Ziegler, Daniel and Wu, Jeffrey and Winter, Clemens and Hesse, Chris and Chen, Mark and Sigler, Eric and Litwin, Mateusz and Gray, Scott and Chess, Benjamin and Clark, Jack and Berner, Christopher and McCandlish, Sam and Radford, Alec and Sutskever, Ilya and Amodei, Dario},
 booktitle = {Advances in Neural Information Processing Systems},
 editor = {H. Larochelle and M. Ranzato and R. Hadsell and M.F. Balcan and H. Lin},
 pages = {1877--1901},
 publisher = {Curran Associates, Inc.},
 title = {Language Models are Few-Shot Learners},
 url = {https://proceedings.neurips.cc/paper_files/paper/2020/file/1457c0d6bfcb4967418bfb8ac142f64a-Paper.pdf},
 volume = {33},
 year = {2020}
}

@inproceedings{zhang-etal-2018-personachat,
    title = "Personalizing Dialogue Agents: {I} have a dog, do you have pets too?",
    author = "Zhang, Saizheng  and
      Dinan, Emily  and
      Urbanek, Jack  and
      Szlam, Arthur  and
      Kiela, Douwe  and
      Weston, Jason",
    editor = "Gurevych, Iryna  and
      Miyao, Yusuke",
    booktitle = "Proceedings of the 56th Annual Meeting of the Association for Computational Linguistics (Volume 1: Long Papers)",
    month = jul,
    year = "2018",
    address = "Melbourne, Australia",
    publisher = "Association for Computational Linguistics",
    url = "https://aclanthology.org/P18-1205",
    doi = "10.18653/v1/P18-1205",
    pages = "2204--2213",
}

@inproceedings{li-etal-2016-persona,
    title = "A Persona-Based Neural Conversation Model",
    author = "Li, Jiwei  and
      Galley, Michel  and
      Brockett, Chris  and
      Spithourakis, Georgios  and
      Gao, Jianfeng  and
      Dolan, Bill",
    editor = "Erk, Katrin  and
      Smith, Noah A.",
    booktitle = "Proceedings of the 54th Annual Meeting of the Association for Computational Linguistics (Volume 1: Long Papers)",
    month = aug,
    year = "2016",
    address = "Berlin, Germany",
    publisher = "Association for Computational Linguistics",
    url = "https://aclanthology.org/P16-1094",
    doi = "10.18653/v1/P16-1094",
    pages = "994--1003",
}

@inproceedings{park-etal-2022-simulacra,
author = {Park, Joon Sung and Popowski, Lindsay and Cai, Carrie and Morris, Meredith Ringel and Liang, Percy and Bernstein, Michael S.},
title = {Social Simulacra: Creating Populated Prototypes for Social Computing Systems},
year = {2022},
isbn = {9781450393201},
publisher = {Association for Computing Machinery},
address = {New York, NY, USA},
url = {https://doi.org/10.1145/3526113.3545616},
doi = {10.1145/3526113.3545616},
booktitle = {Proceedings of the 35th Annual ACM Symposium on User Interface Software and Technology},
articleno = {74},
numpages = {18},
location = {Bend, OR, USA},
series = {UIST '22}
}

@inproceedings{shao-etal-2023-characterllm,
    title = "Character-{LLM}: A Trainable Agent for Role-Playing",
    author = "Shao, Yunfan  and
      Li, Linyang  and
      Dai, Junqi  and
      Qiu, Xipeng",
    editor = "Bouamor, Houda  and
      Pino, Juan  and
      Bali, Kalika",
    booktitle = "Proceedings of the 2023 Conference on Empirical Methods in Natural Language Processing",
    month = dec,
    year = "2023",
    address = "Singapore",
    publisher = "Association for Computational Linguistics",
    url = "https://aclanthology.org/2023.emnlp-main.814",
    pages = "13153--13187",
}

@inproceedings{kano-etal-2023-aiwolfdial,
    title = "{AIW}olf{D}ial 2023: Summary of Natural Language Division of 5th International {AIW}olf Contest",
    author = "Kano, Yoshinobu  and
      Watanabe, Neo  and
      Kagaminuma, Kaito  and
      Aranha, Claus  and
      Lee, Jaewon  and
      Hauer, Benedek  and
      Shibata, Hisaichi  and
      Miki, Soichiro  and
      Nakamura, Yuta  and
      Okubo, Takuya  and
      Shigemura, Soga  and
      Ito, Rei  and
      Takashima, Kazuki  and
      Fukuda, Tomoki  and
      Wakutani, Masahiro  and
      Hatanaka, Tomoya  and
      Uchida, Mami  and
      Abe, Mikio  and
      Mikami, Akihiro  and
      Otsuki, Takashi  and
      Qi, Zhiyang  and
      Harada, Kei  and
      Inaba, Michimasa  and
      Katagami, Daisuke  and
      Osawa, Hirotaka  and
      Toriumi, Fujio",
    editor = "Mille, Simon",
    booktitle = "Proceedings of the 16th International Natural Language Generation Conference: Generation Challenges",
    month = sep,
    year = "2023",
    address = "Prague, Czechia",
    publisher = "Association for Computational Linguistics",
    url = "https://aclanthology.org/2023.inlg-genchal.13",
    pages = "84--100",
}

@article{achiam2023gpt,
  title={GPT-4 Technical Report},
  author={OpenAI},
  journal={arXiv preprint arXiv:2303.08774},
  year={2023}
}

@article{xu2023exploring,
  title={Exploring large language models for communication games: An empirical study on werewolf},
  author={Xu, Yuzhuang and Wang, Shuo and Li, Peng and Luo, Fuwen and Wang, Xiaolong and Liu, Weidong and Liu, Yang},
  journal={arXiv preprint arXiv:2309.04658},
  year={2023}
}

@article{wu2024enhance,
  title={Enhance reasoning for large language models in the game werewolf},
  author={Wu, Shuang and Zhu, Liwen and Yang, Tao and Xu, Shiwei and Fu, Qiang and Wei, Yang and Fu, Haobo},
  journal={arXiv preprint arXiv:2402.02330},
  year={2024}
}

%\appendix

%\section{Example Appendix}
%\label{sec:appendix}

%This is an appendix.

\end{document}